\documentclass{article}

\usepackage{microtype}
\usepackage{graphicx}
\usepackage{comment}
\usepackage{booktabs} 

\usepackage{appendix}
\usepackage{amsmath,amssymb}
\usepackage{subcaption}

\usepackage{hyperref}

\usepackage[accepted]{icml2018}

\icmltitlerunning{Generating Realistic Geology Conditioned on Measurements with GANs}

\begin{document}

\twocolumn[
\icmltitle{Generating Realistic Geology Conditioned on Physical Measurements \\ with Generative Adversarial Networks}



\icmlsetsymbol{}{}

\begin{icmlauthorlist}
\icmlauthor{Emilien Dupont}{stic}
\icmlauthor{Tuanfeng Zhang}{sdr}
\icmlauthor{Peter Tilke}{sdr}
\icmlauthor{Lin Liang}{sdr}
\icmlauthor{William Bailey}{sdr}
\end{icmlauthorlist}

\icmlaffiliation{stic}{Schlumberger Software Technology Innovation Center, Menlo Park, CA, USA}
\icmlaffiliation{sdr}{Schlumberger Doll Research, Cambridge, MA, USA}

\icmlcorrespondingauthor{Tuanfeng Zhang}{tzhang2@slb.com}
\icmlcorrespondingauthor{Peter Tilke}{tilke@slb.com}

\icmlkeywords{Machine Learning, ICML}

\vskip 0.3in
]


\printAffiliationsAndNotice{}  

\begin{abstract}
An important problem in geostatistics is to build models of the subsurface of the Earth given physical measurements at sparse spatial locations. Typically, this is done using spatial interpolation methods or by reproducing patterns from a reference image. However, these algorithms fail to produce realistic patterns and do not exhibit the wide range of uncertainty inherent in the prediction of geology. In this paper, we show how semantic inpainting with Generative Adversarial Networks can be used to generate varied realizations of geology which honor physical measurements while matching the expected geological patterns. In contrast to other algorithms, our method scales well with the number of data points and mimics a \textit{distribution} of patterns as opposed to a single pattern or image. The generated conditional samples are state of the art.
\end{abstract}

\section{Introduction}
\begin{figure}[t]
\begin{center}
\includegraphics[width=0.8\linewidth, clip=true]{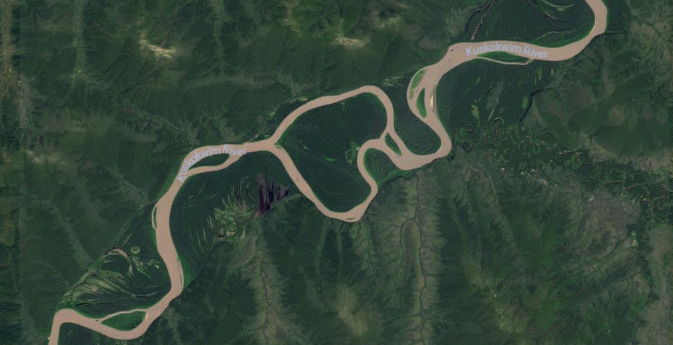}
\end{center}
\caption{Example of non linear meandering fluvial pattern, Kuskokwim River, Alaska.}
\label{river-figure}
\end{figure}

Modeling geology based on sparse physical measurements is key for several use cases including natural resource management, natural hazard identification and geological process quantification. The modeling process involves inferring geological properties, such as rock type, over a wide area given measurements at only a small number of locations. Several tools exist for creating geological models, one of the most important being geostatistics (\citet{deutsch1992geostatistical}). Early geostatistics algorithms mainly used spatial linear interpolation or performed simulation of geological attributes by assuming these attributes follow Gaussian distributions (\citet{cressie1990origins}). Most geological patterns, however, are non Gaussian and highly non linear (Fig. \ref{river-figure}). To overcome these limitations, geostatistical approaches have been developed to simulate complex geological patterns based on a training image (\citet{strebelle2002conditional,zhang2006filter}). These methods aim to generate geological models by extracting patterns from the training image and anchoring them to physical measurements at known locations. However, these algorithms fail to produce realistic non linear patterns and do not exhibit the variability and uncertainty of geological inference.

In this paper, we propose a novel approach, based on deep generative models, for generating realistic geological models that can model a \textit{distribution} of images and reproduce geological patterns while having the flexibility to honor physical measurements. Despite the recent progress in generative modeling, there have only been few examples of geological applications, all of which aim to generate data in an unconstrained fashion (\citet{mosser2017reconstruction,mosser2017stochastic,chan2017parametrization}). By establishing a connection between semantic inpainting and modeling geology, we show how we can generate realistic geological realizations constrained by physical measurements at various spatial locations. To the best of our knowledge, this is the first time the connection between semantic inpainting algorithms and geomodelling has been made. We hope this opens an avenue for generating realistic subsurface geological models constrained by a variety of physical measurements in a seamless fashion. 

While the problem we study is one of geology, we believe our approach is applicable to many fields. Specifically, several tasks in spatio-temporal statistics (\citet{cressie2015statistics}) could benefit from our method, with applications to energy, climate, the environment and agriculture. In general, our algorithm can be used to solve the task of inferring geospatial patterns of quantities such as temperature, pressure, humidity and porosity from sparse measurements. As an example of an alternative use case we have applied our algorithm to satellite images of river deltas. 

\section{Method}
\subsection{Problem Statement}
Given a number of true physical measurements and a distribution of geological patterns (e.g. a large set of images), we would like to generate realizations of the geology which honor the physical measurements while producing realistic geological patterns. More specifically, suppose that a specific area is known to have, for example, fluvial geological patterns (see Fig.~\ref{river-figure}). Suppose also that, at various spatial locations $(x, y)$, the type of the rock $r$ has been measured, i.e. we have set of tuples of the form $(x, y, r)$. Given these measurements, we wish to infer the rock type $r$ at all locations such that:
\begin{enumerate}
  \item The resulting realizations are realistic (i.e. match the distribution of geological patterns)
  \item The resulting realizations honor the physical measurements
\end{enumerate}

\begin{figure}[t]
\begin{center}
\includegraphics[width=0.8\linewidth]{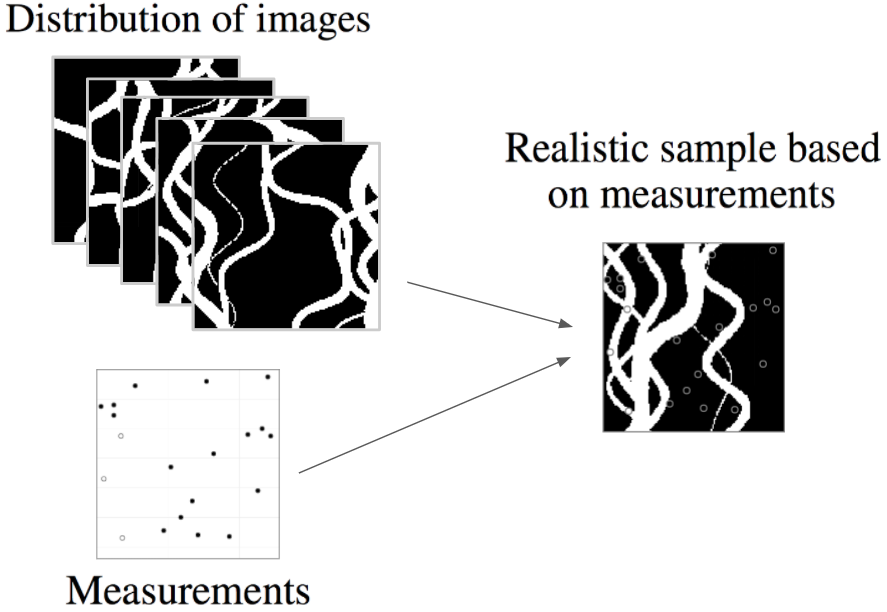}
\end{center}
\caption{Illustration of the geological problem. Given a set of sparse measurements of rock types in an area containing known geological patterns, infer rock type at all locations.}
\label{explanation-sketch}
\end{figure}

In this paper we restrict ourselves to the case of fluvial patterns with binary measurements (i.e. there are only two types of rock present), although it is easy to extend this to the general case. Specifically, we have a distribution of binary images $p_{fluvial}(z)$ where $z \in\{0, 1\}^{n \times n}$ and a set of $m$ binary measurements $(x, y, r)$ where $r \in \{0, 1\}$. Our goal is to generate samples $z \sim p_{fluvial}(z)$ such that $z(x_i, y_i) = r_i$ for $i=1,...,m$. This approach is illustrated in Fig. \ref{explanation-sketch}.

\subsection{Combining Semantic Inpainting and Geology}

The above problem is closely related to semantic inpainting. Semantic inpainting is the task of filling in missing regions of an image using the surrounding pixels and a prior on what the images should look like. Recent semantic inpainting algorithms have taken advantage of deep generative models to create realistic infillings of various types of images (\citet{pathak2016context,yeh2016semantic,li2017context}). \citet{yeh2016semantic} in particular propose a framework for semantic inpainting based on Generative Adversarial Networks (GAN) (\citet{goodfellow2014generative}). GANs are generative models composed of a generator $G$ and discriminator $D$ each parametrized by a neural network. $G$ is trained to map a latent vector $\mathbf{z}$ into an image $\mathbf{x}$, while $D$ is trained to map an image $\mathbf{x}$ to the probability of it being real (as opposed to generated). The networks are trained adversarially by optimizing the loss

$$\min_{G} \max_{D} \mathbb{E}_{\mathbf{x} \sim p_{data}(\mathbf{x})} [\log D(\mathbf{x})] + \mathbb{E}_{\mathbf{z} \sim p(\mathbf{z})} [\log (1 - D(G(\mathbf{z})))]    $$

After training on data drawn from a distribution $p_{data}$, $G$ will be able to generate samples similar to those from $p_{data}$, by sampling $\mathbf{z} \sim p(\mathbf{z})$ and mapping $\mathbf{x} = G(\mathbf{z})$. Using a trained $G$ and $D$, we would like to generate realistic images $\mathbf{x}_g=G(\mathbf{z})$ conditioned on a set of known pixel values $\mathbf{y}$. This can be achieved by fixing the weights of $G$ and $D$ and optimizing $\mathbf{z}$ to generate realistic samples based on the known pixel values. In order to generate realistic samples we would like the samples $\mathbf{x}_g$ to be close to $p_{data}$, i.e. we would like to generate samples such that the discriminator $D$ assigns high probability to $\mathbf{x}_g$. This idea is formalized through the \textit{prior loss}. We would also like the samples to honor the pixel values $\mathbf{y}$, i.e. we want the generated $\mathbf{x}_g$ to match $\mathbf{y}$ at the known pixel locations. This is formalized through the \textit{context loss}.

\begin{figure}[t]
\begin{center}
\includegraphics[width=0.8\linewidth]{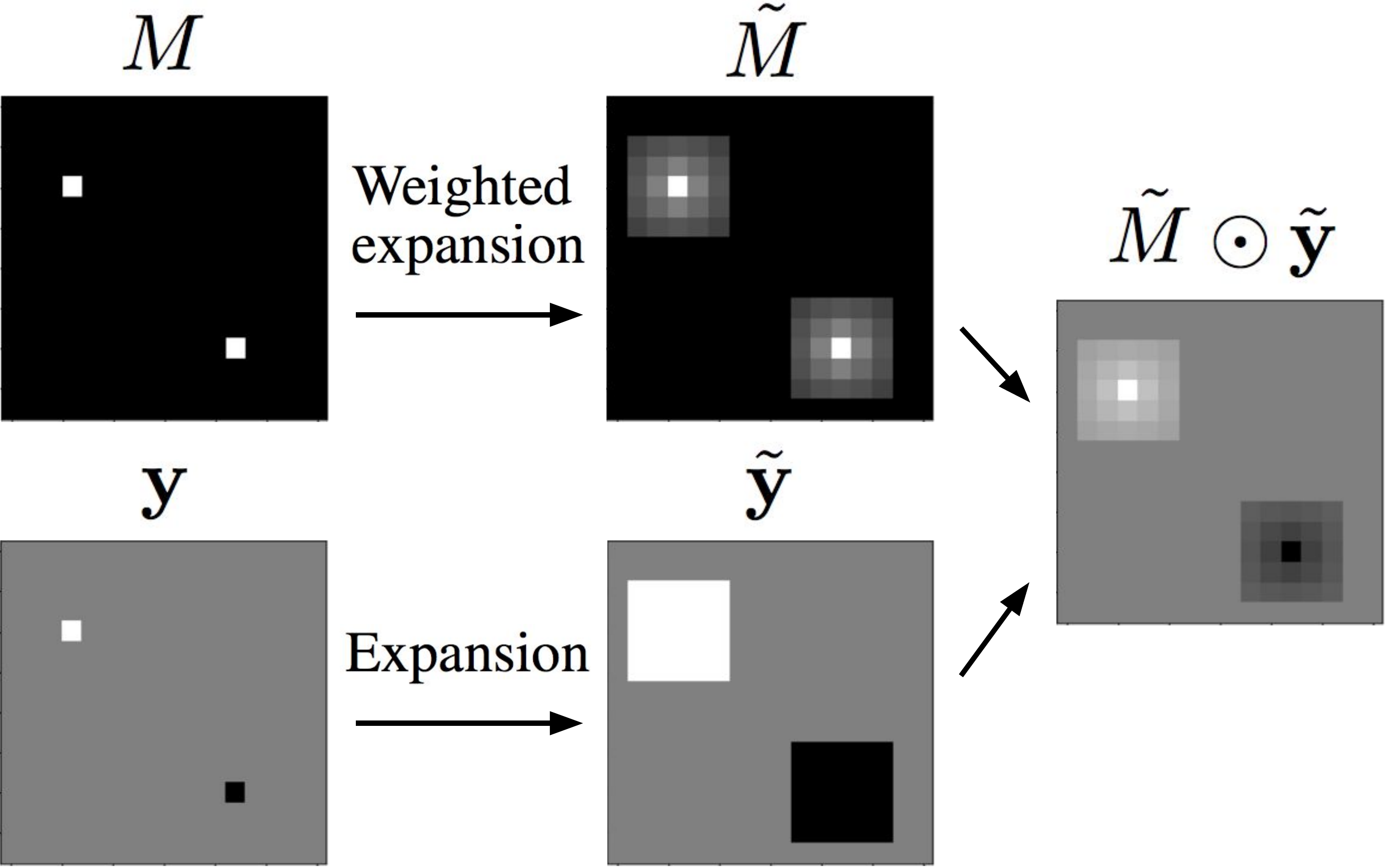}
\end{center}
\caption{Expanding context of known pixels. A mask $M$ with two visible pixels (white denotes a visible pixel, black a hidden pixel) and the corresponding known pixel values $\mathbf{y}$ (white corresponds to a pixel with value 1, black to a pixel with value 0 and gray to an unknown pixel value) are shown on the left. Each visible pixel is expanded in both the mask and the known values (in a weighted and unweighted manner respectively).}
\label{mask-expansion}
\end{figure}

\textbf{Prior loss}: The prior loss $\mathcal{L}_p$ penalizes unrealistic images. Since $D$ was trained to assign high probability to realistic samples, the prior loss is chosen as
$$ \mathcal{L}_p(\mathbf{z}) = \log (1 - D(G(\mathbf{z})))  $$

\textbf{Context loss}: The context loss penalizes mismatch between generated samples and known pixels $\mathbf{y}$. Denoting by $M$ the masking matrix (which has value 1 at the known points and 0 otherwise), the context loss is defined by \citet{yeh2016semantic} as

$$ \mathcal{L}_c(\mathbf{z}|\mathbf{y}, M) = \| M \odot (G(\mathbf{z}) - \mathbf{y})  \|_1 $$

Since the set of known pixels $\mathbf{y}$ is very sparse, we modify the objective by expanding the mask $M$. Specifically, for every unmasked value, i.e. for every $(i,j)$ such that $M_{ij}=1$, we modify the neighboring pixels within a set radius to have value 
$$M_{i+\delta i,j+\delta j}=\frac{1}{\sqrt[]{\delta i^2 + \delta j^2 + 1}}$$ 

which can be thought of as changing the mask from imposing a hard constraint to a smooth constraint. We perform a corresponding expansion for the set of known pixel values $\mathbf{y}$ (see Fig. \ref{mask-expansion}). This encourages the generator to output samples which not only match the known pixels but also produce similar values around the known pixels. This modification was found to facilitate the optimization of $\mathcal{L}_c$, especially when $M$ and $y$ were very sparse. Denoting by $\tilde{M}$ the expanded weighted masking matrix and by $\mathbf{\tilde{y}}$ the expanded pixels, the modified context loss can be written as 

$$ \mathcal{\tilde{L}}_c(\mathbf{z}|\mathbf{y}, M) = \| \tilde{M} \odot (G(\mathbf{z}) - \mathbf{\tilde{y}})  \|_1 $$

\textbf{Total loss} The total loss is defined as the weighted sum of the prior loss and context loss
\begin{equation}
\mathcal{L}(\mathbf{z}) = \mathcal{\tilde{L}}_c(\mathbf{z}|\mathbf{y}, M) + \lambda \mathcal{L}_p(\mathbf{z}) 
\label{eq:1}
\end{equation}

where $\lambda$ controls the tradeoff between generating realistic images and matching the known pixel values.

We can now establish a connection between semantic inpainting and generating realizations of geology conditioned on physical measurements. We let $p_{data}$ be the distribution of fluvial patterns $p_{fluvial}$ and we let the set of known pixel values $\mathbf{y}$ be the set of known physical measurements at various locations. By randomly initializing $\mathbf{z}$ and minimizing $\mathcal{L}(\mathbf{z})$ we can then obtain various realizations of the geology that are realistic while honoring the physical measurements (the relationship between using different initializations of $\mathbf{z}$ and generating different realizations of the geology is discussed in more detail in Section \ref{conditional-model}). The final algorithm for generating a realization conditioned on physical measurements can then be described as follows. Given distribution of fluvial patterns $p_{fluvial}(\mathbf{x})$, measurements $\mathbf{y}$:
\begin{enumerate}
  \item Train $G$ and $D$ on $p_{fluvial}(\mathbf{x})$
  \item Initialize $\mathbf{z}$ randomly
  \item Compute $\mathbf{z}^*$ by minimizing $\mathcal{L}(\mathbf{z})$
  \item Return sample $\mathbf{x} = G(\mathbf{z}^*)$
\end{enumerate}

\begin{figure}[h]
\begin{center}
\includegraphics[width=1.0\linewidth]{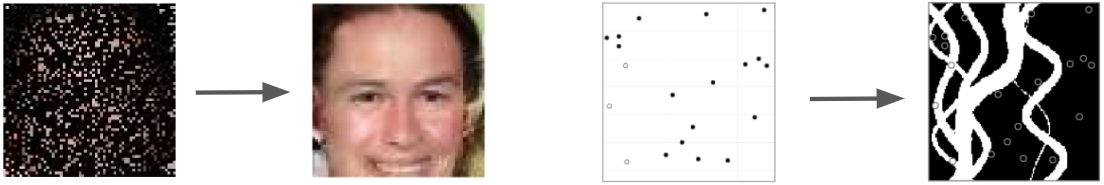}
\end{center}
\caption{Comparison between inferring missing pixel values and generating geology based on physical measurements. Image on the left borrowed from \citet{yeh2016semantic}}.
\label{semantic-comparison}
\end{figure}

\subsection{Generating Data}

To train $G$ and $D$ we require a large dataset of geological patterns. In order to create such a dataset we take advantage of another family of geomodelling approaches called object-based models (OBM). OBMs are capable of generating very realistic geobodies based on distributions of geometric shapes using marked point processes (\citet{deutsch1996hierarchical}). Note that these models can also be conditioned to physical measurements using Markov Chain Monte Carlo (MCMC) algorithms (\citet{holden1998modeling}). Even though OBMs are good at producing realistic geobodies, conditioning on dense data locations is very difficult and extremely slow and often fails to converge in MCMC (\citet{skorstad1999well, hauge2007well}).

\begin{figure}[t]
\centering
\begin{subfigure}[t]{0.6\linewidth}
\centering
\includegraphics[width=1.0\linewidth]{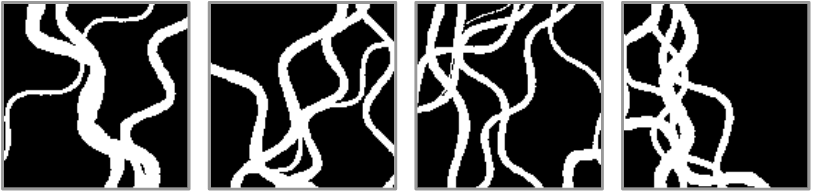}
\caption{Training data.}
\label{training-data}
\end{subfigure}
\begin{subfigure}[t]{0.9\linewidth}
\centering
\includegraphics[width=1.0\linewidth]{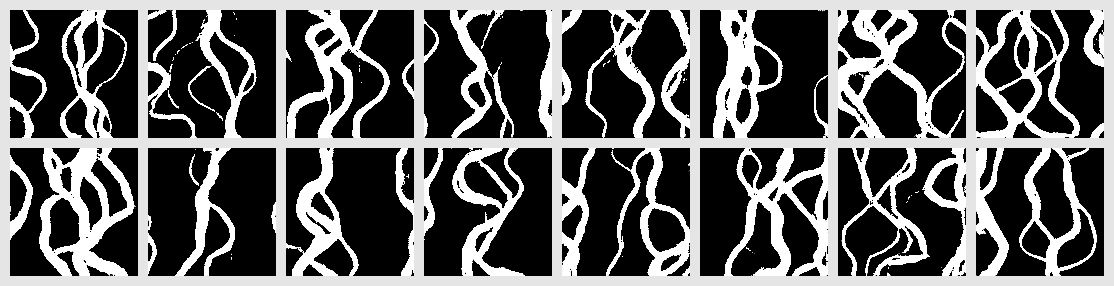}
\caption{Generated data.}
\label{generated-samples}
\end{subfigure}
\caption{(a) Examples of training data (generated with OBM). White pixels correspond to fluvial channels while black pixels correspond to background rock. (b) Generated samples. The samples are sharp and exhibit many of the desired properties such as channel connectivity and variety. For more examples please refer to appendix.}
\end{figure}

We use OBM to generate a training set of 5000 images of fluvial patterns. Each image has dimension 128 by 128 and is binary, with 1 corresponding to channels and 0 corresponding to background (i.e. two different types of rock). More details on data generation are described in the appendix. Examples of the training data are shown in Fig. \ref{training-data}.

\section{Experiments}
\subsection{Unconditional model}
We use a DC-GAN architecture (\citet{radford2015unsupervised}) for the discriminator and generator and train an unconditional model (i.e. a model that freely generates fluvial patterns without being restricted by physical measurements) on the dataset of 5000 images (see appendix for details of model architecture and training). 

\begin{figure}[t]
\centering
\begin{subfigure}[h]{0.2\linewidth}
\centering
\includegraphics[width=1.0\linewidth]{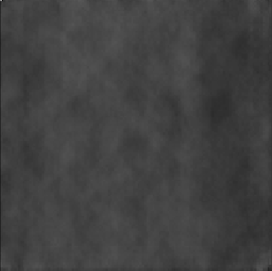}
\caption{}
\label{mean-unconditional}
\end{subfigure}\hspace{0.1\linewidth}
\begin{subfigure}[h]{0.6\linewidth}
\centering
\includegraphics[width=1.0\linewidth]{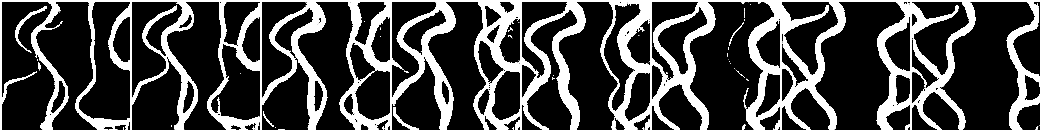}
\includegraphics[width=1.0\linewidth]{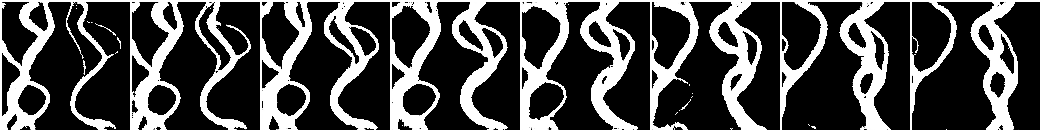}
\includegraphics[width=1.0\linewidth]{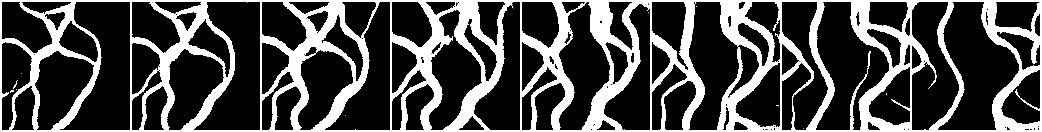}
\includegraphics[width=1.0\linewidth]{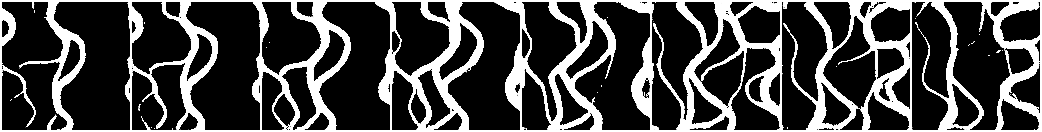}
\caption{}
\label{latent-traversal}
\end{subfigure}
\caption{(a) Mean image of a 1000 unconditional samples. The pixel distribution is fairly uniform suggesting the model has learnt an even distribution of channels. (b) Examples of latent traversals. Each row of images was obtained by interpolating between two random vectors $\mathbf{z}_1$ and $\mathbf{z}_2$ drawn from $\mathcal{N}(0, 2)$ and mapping them through $G$. }
\end{figure}

In order to verify that the model has learnt to generate realistic and varied samples, we perform several checks. Firstly, we visually inspect samples from the model (see Fig. \ref{generated-samples}). The samples are almost indistinguishable from the true data and embody a lot of the properties of the data, e.g. the channels remain connected, the patterns are meandering and there is a large diversity suggesting the model has learnt a wide distribution of the data. Secondly, since the training images were generated by OBM, we know some of the statistics the model should satisfy (see appendix). For example, the ratio of white pixels to black pixels is a generative factor (and was chosen to be 0.25). Sampling a 1000 images from $G$, and calculating the ratio of white to black pixels we obtain the same ratio within 1\%. In addition, to ensure the channels are well distributed, we average a 1000 generated images and check that the pixel values are approximately 0.25 everywhere (see Fig. \ref{mean-unconditional}). Finally, we can also perform traversals in latent space to ensure the model has learnt a good approximation of the data manifold. We would for example expect the changes over the latent space to be smooth and not disconnect any channels. This is shown in Fig. \ref{latent-traversal}.

\subsection{Conditional model}
\label{conditional-model}
\begin{figure}[t]
\begin{center}
\includegraphics[width=1.0\linewidth]{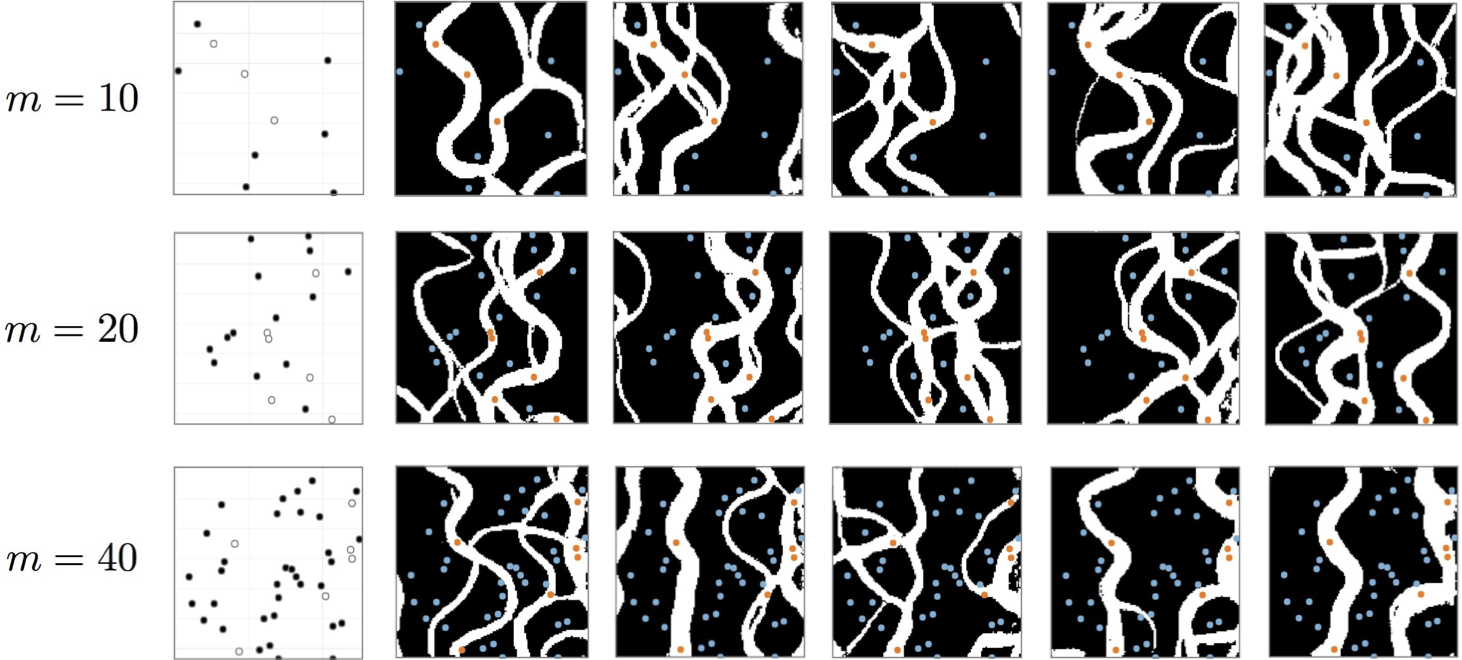}
\end{center}
\caption{Conditional samples. Each row corresponds to a different number of measurements. The first column shows the true measurements while the subsequent columns show different realizations based on the measurements (we show the 5 best realizations chosen by visual inspection from 20 runs). The realizations are realistic and honor the data. Note that the measurements are shown in blue and orange on the generated samples for clarity.}
\label{conditioned-samples}
\end{figure}

\begin{figure}[h]
\begin{center}
\includegraphics[width=0.8\linewidth]{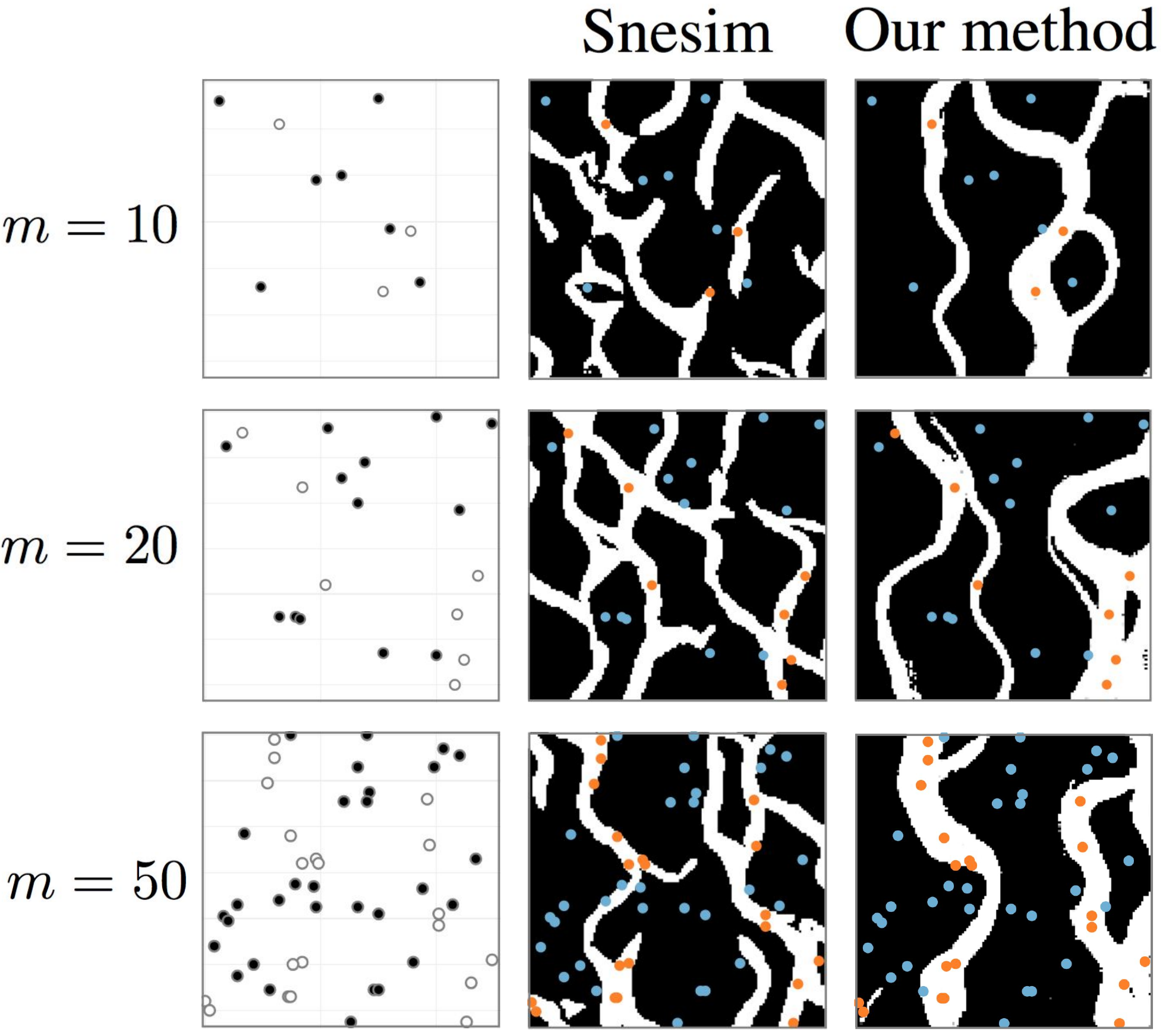}
\end{center}
\caption{Comparison between snesim (which is described in Section \ref{related-work}) and our approach. The left column shows the sparse measurements which were used for both snesim and our method. The middle column shows samples generated by snesim (best realization chosen by visual inspection from 10 runs) and the right column shows samples generated by our method (best of 5 runs). While both algorithms honor the data, snesim produces disconnected and unrealistic channels, while our method generates realistic and coherent samples.}
\label{snesim-comparison}
\end{figure}

Using the unconditional GAN, we can create samples conditioned on measurements by minimizing the loss $\mathcal{L}(\mathbf{z})$ specified in equation (\ref{eq:1}). To do this, we take a fluvial image and randomly select $m$ measurement locations and mask the rest of the input. We then randomly initialize several $\mathbf{z}^{(i)}$ and minimize $\mathcal{L}(\mathbf{z})$ for each of them using Adam (\citet{kingma2014adam}). This results in a set of $\mathbf{z}^{*(i)}$ each corresponding to a local minimum of $\mathcal{L}(\mathbf{z})$. Each of the $\mathbf{z}^{*(i)}$ is mapped to a realization $\mathbf{x}^{(i)}=G(\mathbf{z}^{*(i)})$ such that the $\mathbf{x}^{(i)}$ honor the physical measurements, have realistic geological patterns and each correspond to \textit{different} realizations. The idea is that different initializations will lead to different minima of $\mathcal{L}(\mathbf{z})$ and in turn a variety of samples all satisfying the constraints. Examples of this are shown in Fig. \ref{conditioned-samples}. As can be seen, the samples produced have high quality (the channels remain connected, the images look realistic) and satisfy the constraints while being fairly diverse. The samples produced are at least comparable to the current state of the art (see Fig. \ref{snesim-comparison}). Further, since the model captures patterns from a distribution of images as opposed to a single training image, the generated samples exhibit a variety of channel widths and curvatures which current algorithms cannot achieve.

\subsubsection{Large number of measurements}

For most geostatistical modeling algorithms, increasing the number of measurements to condition on, degrades performance and increases runtime (\citet{strebelle2002conditional,zhang2006filter}). In particular, the generated images often become unrealistic as more data constraints are added. This is not the case for our algorithm and we are able to generate realistic patterns over a wide number of measurements from sparse to dense. Results are shown in Fig. \ref{dense-conditioning}. 

\subsubsection{Failure cases}

\begin{figure}[t]
\centering
\begin{subfigure}[t]{0.7\linewidth}
\centering
\includegraphics[width=1.0\linewidth]{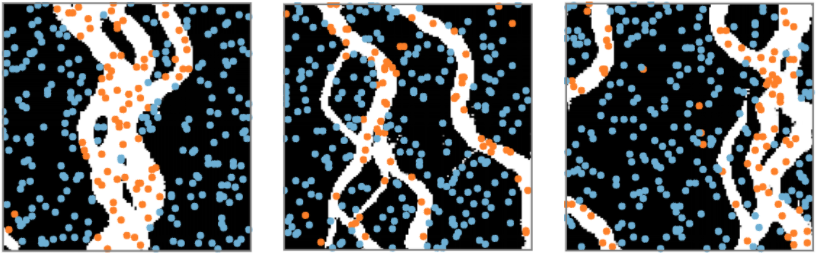}
\caption{}
\label{dense-conditioning}
\end{subfigure}
\begin{subfigure}[t]{0.8\linewidth}
\centering
\includegraphics[width=1.0\linewidth]{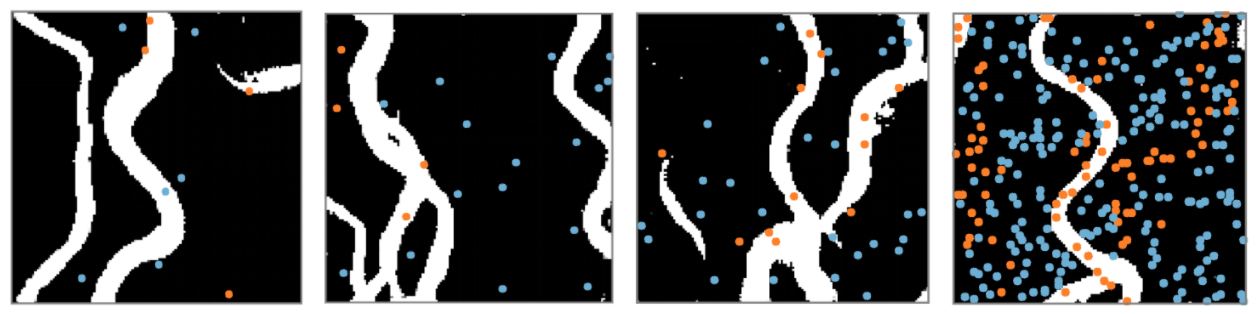}
\caption{}
\label{failure-examples}
\end{subfigure}
\caption{(a) Samples conditioned on 300 points. Even with a large number of physical measurements the algorithm generates realistic realizations while honoring most data points. (b) Failure examples for different number of measurements ($m$=10, 20, 40, 300). The samples either fail to honor the measurements or produce unrealistic patterns (e.g. blobs of disconnected channels).}
\end{figure}

Minimizing $\mathcal{L}(\mathbf{z})$ can sometimes lead to poor minima and in turn samples that are either unrealistic or do not satisfy the measurement constraints. We have included examples of this in Fig. \ref{failure-examples}.

\subsubsection{Sample diversity}

While the unconditional model exhibits large sample diversity, this is not always the case for the conditional samples. The conditional samples created are realistic and honor the physical measurements but may lack diversity. The variety of samples is based on the variety of local minima of $\mathcal{L}(\mathbf{z})$ and random initializations of $\mathbf{z}$ may end up in the same local minimum, resulting in the same sample. It would be interesting to build semantic inpainting algorithms based on true sampling and not on solving an optimization problem. To the best of our knowledge, creating models that can simultaneously (a) generate realistic images, (b) honor constraints, (c) exhibit high sample diversity is an open problem.

\subsubsection{Application to Satellite Images}

\begin{figure}[t]
\centering
\begin{subfigure}[h]{0.4\linewidth}
\centering
\includegraphics[width=1.0\linewidth]{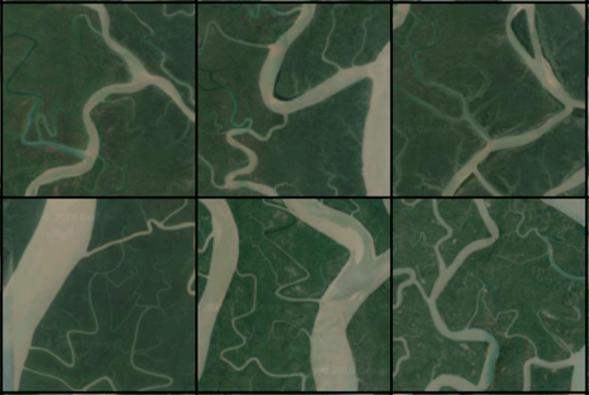}
\caption{Real samples.}
\label{satellite-real}
\end{subfigure}\hspace{0.1\linewidth}
\begin{subfigure}[h]{0.4\linewidth}
\centering
\includegraphics[width=1.0\linewidth]{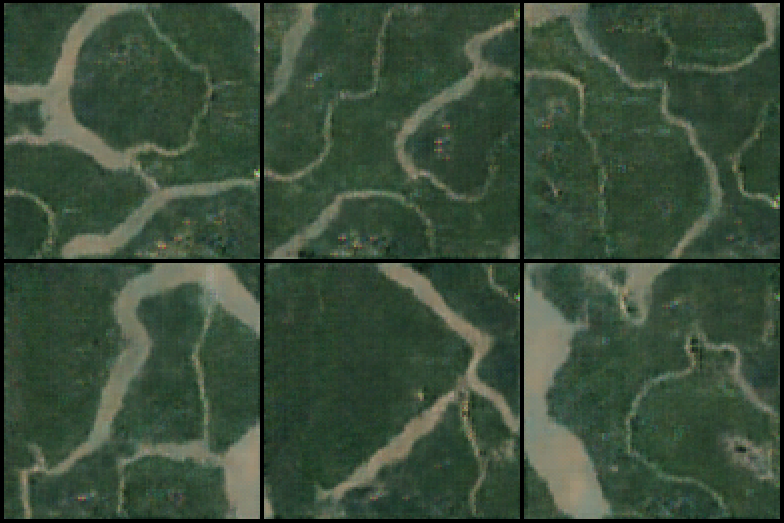}
\caption{Generated samples.}
\label{satellite-fake}
\end{subfigure}
\caption{Comparison of real and generated samples.}
\end{figure}

\begin{figure}[b]
\begin{center}
\includegraphics[width=1.0\linewidth]{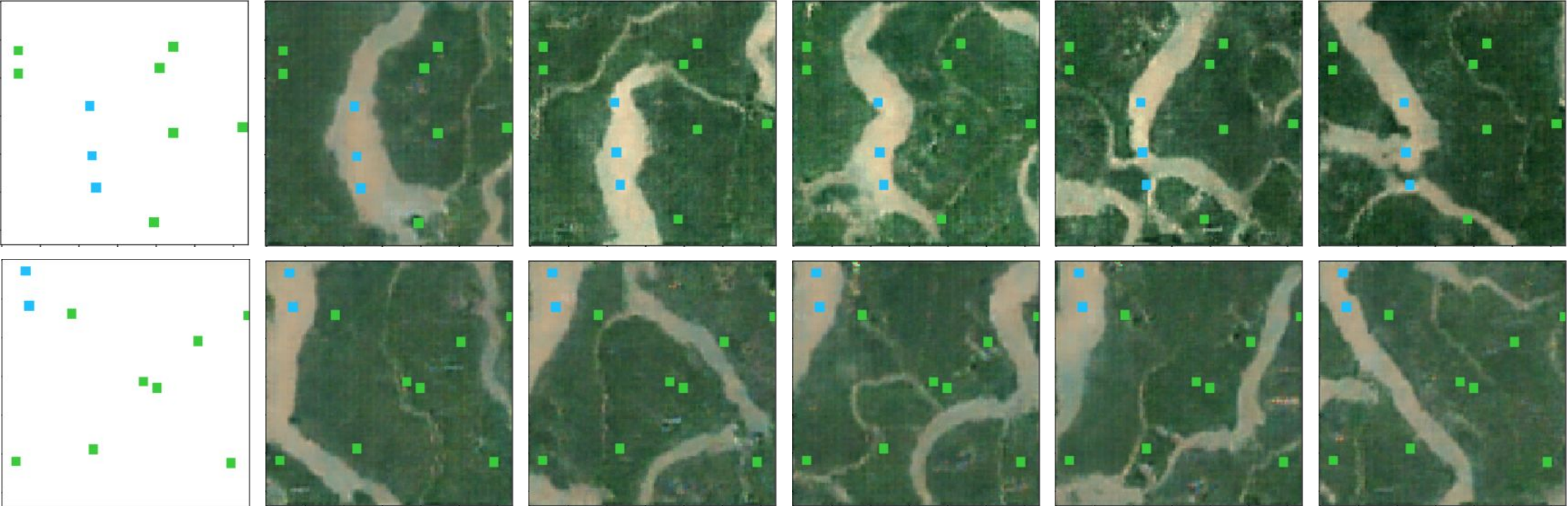}
\end{center}
\caption{Samples conditioned on measurements. The first column shows the conditioning points (blue indicates water and green indicates land) and the other columns show realizations generated by our method. As can be seen the samples are realistic and diverse while honoring the conditioning.}
\label{satellite-conditional}
\end{figure}

To demonstrate the wide applicability of our method we also trained a model on satellite images from the Landsat dataset (\citet{usgs2018landsat}). We collected 3000 satellite images from the Ganges delta in Bangladesh and India, which is a region containing a wide variety of fluvial patterns. We trained the same model that was used for the geological application on 128 by 128 versions of these satellite images. Examples of real and generated images are shown in Fig. \ref{satellite-real} and \ref{satellite-fake}.

We then randomly fixed the location of a number of known measurements. For example, a certain latitude and longitude would correspond to a point containing water and another to a point containing land. In a similar fashion to the geological application, we then minimized the loss $\mathcal{L}(\mathbf{z})$ to match the generated patterns with the known measurements. Results are shown in Fig. \ref{satellite-conditional}. As can be seen, the results are convincing: the generated samples look realistic while honoring the sparse measurement constraints.

While we have applied our method to generate constrained geological and satellite data, we believe the applicability of our method goes beyond this and we hope it will be used on other types of geospatial data in fields such as energy, climate, environment and agriculture.

\section{Related Work}
\label{related-work}
\citet{strebelle2002conditional} proposed an algorithm, called snesim, based on a single training image. The algorithm works by scanning a training image and storing all the training patterns in a search tree. The simulation then proceeds by reproducing the complex patterns from the training image by inferring a local probability density function from scanning the search tree.  

\citet{zhang2006filter} proposed another approach based on filters. The algorithm simulates patterns from a training image by matching them with filters. A fixed set of filters is used to scan the training image, classifying training patterns by their filter scores (corresponding to convolutions of the filter with a patch of the training image). Each filter sees different aspects of the patterns in the image, e.g. an average filter for pattern location, a gradient filter for boundary detection and a curvature filter for second order changes. The simulation then proceeds in a sequential manner by visiting each pixel randomly and patching the patterns closest to the training image while honoring the previously simulated patches and measurements.

These algorithms have the flexibility of honoring measurements while reasonably capturing patterns from a training image. However, it is impractical to use these algorithms with several training images and so the simulations fail to correctly reflect the inherent uncertainty of geological predictions. Additionally, the generated samples are often not realistic, for example containing broken channels even when the training image contains only connected channels.

\citet{chan2017parametrization} propose an approach for parameterizing a geological model using GANs. The authors use a single training image and split it into smaller patches to be used as a training set. In contrast, we use several training images with wide uncertainty and generate large coherent realizations of the geology as opposed to small patches. More importantly, our approach allows for conditioning on physical measurements. The ability to condition on measurements while generating realistic work is the key contribution of our work since other methods, such as OBM, already exist for generating unconditioned samples (\citet{deutsch1996hierarchical}).

\section{Conclusion}
We have proposed a powerful and flexible framework for generating realizations of geology conditioned on physical measurements. It is superior to existing geological modeling tools in several aspects. Firstly, it is able to generate realistic geological realizations with a wide range of uncertainty by capturing a distribution of patterns as opposed to a single pattern or image. Furthermore, it is able to condition on a much larger number of physical measurements than previous algorithms, while still generating realistic samples. 

In future work we plan to expand this to generate subsurface geological models in 3D and to include more than two types of rock. While we have applied this framework to geostatistics, we believe that such an approach could be useful for other scientific problems where sparse measurements and a prior on patterns are available.

\subsubsection*{Acknowledgments}

The authors would like to thank Erik Burton, Jos\'{e} Celaya, Suhas Suresha and Vishakh Hegde for helpful suggestions and comments.

\bibliographystyle{ACM-Reference-Format}
\bibliography{iclr2018_workshop}

\appendix

\section{Generated samples}

\begin{figure}[H]
\begin{center}
\includegraphics[width=0.95\linewidth]{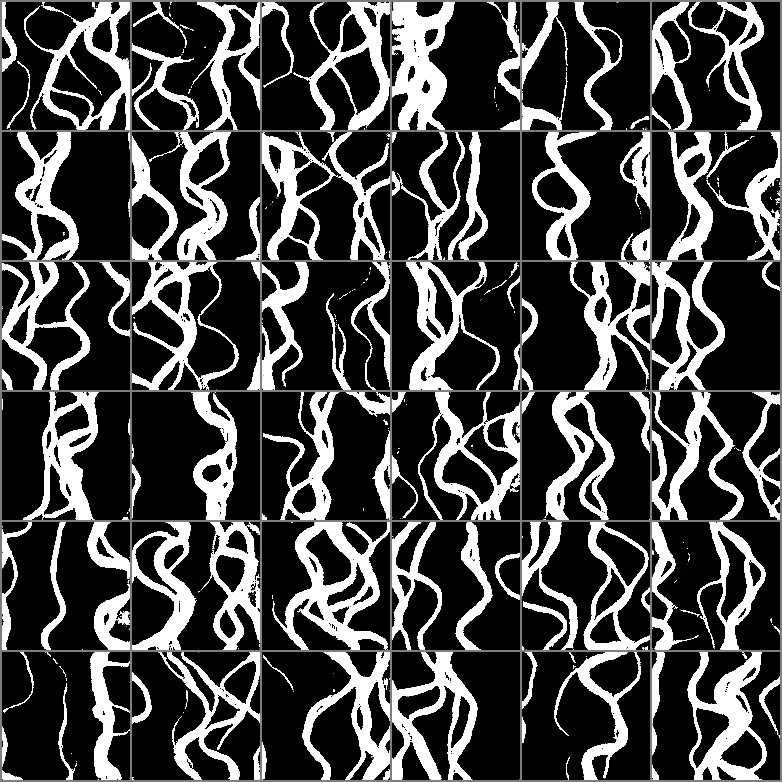}
\end{center}
\end{figure}

\begin{figure}[H]
\begin{center}
\includegraphics[width=0.95\linewidth]{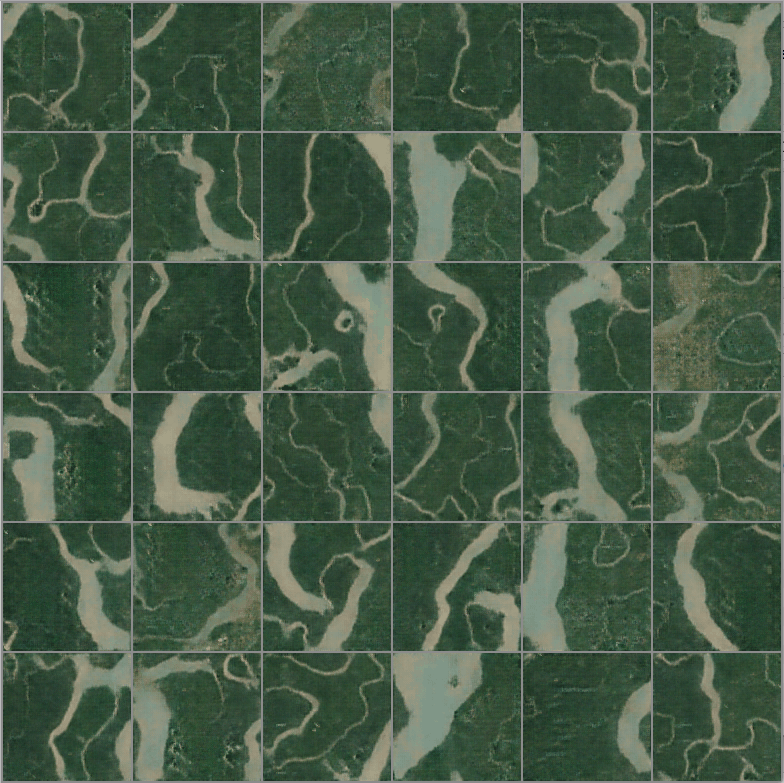}
\end{center}
\end{figure}

\section{Training data generation}
The data was generated with OBM using three channel parameters: orientation, amplitude and wavelength. Each of these follow a triangular distribution. The orientation was varied from -60 to 60 degrees; the amplitude from 10 to 30; and the wavelength from 50 to 100. The channel proportion for each image was controlled to be 25\% (i.e. on average 25\% of the pixels are white). See \citet{deutsch1996hierarchical} for further details on OBM.

\section{Model and training}

The architecture of the model is shown in the below table. The non linearities in the discriminator are LeakyReLU(0.2) except for the output layer which is a sigmoid. The non linearities in the generator are ReLU except for the output layer which is a tanh. The models were trained for 500 epochs with Adam with a learning rate of 1e-4, $\beta_1=0$ and $\beta_2=0.9$.
\begin{table}[h]
\begin{center}
\begin{tabular}{c c}
\textbf{Discriminator} & \textbf{Generator}
\\ \hline \\
64 Conv $4 \times 4$, stride 2        & Linear $100 \times 2048$   \\
128 Conv $4 \times 4$, stride 2       & 256 Conv Transpose $4 \times 4$, stride 2   \\
256 Conv $4 \times 4$, stride 2       & 128 Conv Transpose $4 \times 4$, stride 2\\
32 Conv $4 \times 4$, stride 2        & 64 Conv Transpose $4 \times 4$, stride 2 \\
Linear $2048 \times 1$				  & 1 Conv Transpose $4 \times 4$, stride 2 \\
\end{tabular}
\end{center}
\end{table}

When optimizing $\mathbf{z}$ to honor the conditional pixels, we used Adam with a learning rate of 1e-2 and default $\beta$ parameters. We used $\lambda=10$ and trained for 1500 iterations. The expansion radius of the mask was varied between 1 and 10 depending on the density of the measurements and the model was found to be fairly robust to variation of this radius. In our experiments, we used a radius of 10 for $m=10$, a radius of 7 for $m=20$, a radius of 5 for $m=40, 50$ and a radius of 1 for $m=300$.

\end{document}